\documentclass[conference]{IEEEtran}
\IEEEoverridecommandlockouts
\usepackage{cite}
\usepackage{amsmath,amssymb,amsfonts}
\usepackage{algorithmic}
\usepackage{graphicx}
\usepackage{textcomp}
\usepackage{xcolor}
\def\BibTeX{{\rm B\kern-.05em{\sc i\kern-.025em b}\kern-.08em
    T\kern-.1667em\lower.7ex\hbox{E}\kern-.125emX}}
\begin{document}

\title{Research on Milling Machine Predictive Maintenance Based on Machine Learning and SHAP Analysis in Intelligent Manufacturing Environment
}

\author{
\begin{minipage}{\textwidth}
\centering
\begin{tabular}{cc}

\begin{tabular}[t]{@{}c@{}}
Wen Zhao \\
\textit{School of Advanced Manufacturing Engineering} \\
\textit{Chongqing University of Posts and Telecommunications} \\
ChongQing, China \\
wenzhao0211@outlook.com
\end{tabular}
&
\begin{tabular}[t]{@{}c@{}}
Jiawen Ding \\
\textit{School of Computing and Information Systems} \\
\textit{The University of Melbourne} \\
Melbourne, Australia \\
jiawending@student.unimelb.edu.au
\end{tabular}
\\[2em]

\begin{tabular}[t]{@{}c@{}}
Xueting Huang \\
\textit{School of Science, Computing and Engineering Technologies} \\
\textit{Swinburne University of Technology} \\
Melbourne, Australia \\
xuetinghuang@swin.edu.au
\end{tabular}
&
\begin{tabular}[t]{@{}c@{}}
Yibo Zhang \\
\textit{Gezhi Future Research Institute} \\
Beijing, China \\
bernie.zhangyibo@gmail.com
\end{tabular}

\end{tabular}
\end{minipage}
}

\maketitle

\begin{abstract}
In the context of intelligent manufacturing, this paper conducts a series of experimental studies on the predictive maintenance of industrial milling machine equipment based on the AI4I 2020 dataset. This paper proposes a complete predictive maintenance experimental process combining artificial intelligence technology, including six main links: data preprocessing, model training, model evaluation, model selection, SHAP analysis, and result visualization. By comparing and analyzing the performance of eight machine learning models, it is found that integrated learning methods such as XGBoost and random forest perform well in milling machine fault prediction tasks. In addition, with the help of SHAP analysis technology, the influence mechanism of different features on equipment failure is deeply revealed, among which processing temperature, torque and speed are the key factors affecting failure. This study combines artificial intelligence and manufacturing technology, provides a methodological reference for predictive maintenance practice in an intelligent manufacturing environment, and has practical significance for promoting the digital transformation of the manufacturing industry, improving production efficiency and reducing maintenance costs.
\end{abstract}


\section{Introduction}

The emergence of Industry 4.0 has transformed the manufacturing paradigm by integrating cyber-physical systems, the Internet of Things (IoT), and data-driven intelligence into traditional industrial processes. Among the key enablers of this transformation, artificial intelligence (AI) plays a pivotal role in enabling predictive, adaptive, and autonomous decision-making across the production lifecycle. One of the most promising applications of AI in this domain is predictive maintenance (PdM), which aims to forecast equipment failures in advance and schedule maintenance accordingly, thereby reducing unplanned downtimes, optimizing resource allocation, and extending machinery lifespan\cite{b1,b2,b3,b4,b5}.

Traditional maintenance strategies, including reactive maintenance (after failure) and preventive maintenance (at fixed intervals), are increasingly inadequate in high-precision and cost-sensitive manufacturing environments. These methods often result in excessive maintenance costs or unexpected breakdowns, as they fail to adapt to the dynamic operating conditions of modern equipment\cite{b6,b7,b8,b9}. Consequently, the demand for intelligent, real-time, and interpretable PdM systems has grown rapidly. In particular, machine learning (ML) algorithms trained on operational data can model complex nonlinear patterns and anticipate potential faults more accurately than rule-based methods\cite{b10,b11,b12}.

Despite substantial progress, two major challenges remain unresolved: (1) Many existing studies prioritize model accuracy but overlook interpretability, limiting trust and practical adoption by domain experts; (2) The comparative performance of diverse ML algorithms in real-world fault prediction tasks—especially for multi-fault classification on high-value equipment like milling machines—is not thoroughly examined in current literature.

To bridge these gaps, this study proposes an end-to-end PdM framework tailored for industrial milling machines using the AI4I 2020 dataset\cite{b16}. The framework involves six key phases: data preprocessing, multi-model training, performance evaluation, hyperparameter optimization, SHAP-based interpretability analysis, and result visualization. To enhance transparency, we incorporate SHAP (SHapley Additive exPlanations) for both global and local feature attribution, enabling clear identification of the operational variables (e.g., temperature, torque, speed) that contribute to failure risk\cite{b17,b18,b19}.

The main contributions of this paper are threefold:
\begin{itemize}
    \item A systematic benchmarking of eight machine learning algorithms for both binary and multi-label fault prediction of milling machines;
    \item A robust experimental workflow that includes hyperparameter tuning, cross-validation, and visual diagnostics to ensure model reliability;
    \item The integration of SHAP analysis to interpret model decisions, reveal critical failure-driving features, and provide practical guidance for maintenance planning.
\end{itemize}

This research provides new evidence on the feasibility and practicality of interpretable AI for industrial predictive maintenance and offers actionable insights that support intelligent decision-making and operational resilience in smart factories\cite{b20,b21,b22,b23,b24,b25}.

\section{Methods}

\subsection{Dataset Introduction}

This study uses the AI4I 2020 Predictive Maintenance dataset, which is a synthetically generated yet highly realistic dataset simulating the operational characteristics of industrial milling machines. The dataset comprises 10,000 records and 14 features, representing a diverse range of operational conditions and failure scenarios. It was specifically designed to support research in intelligent fault diagnosis and condition monitoring for predictive maintenance systems. The main features include both categorical and numerical variables, such as unique identifier (UID), product ID, product type (categorical), air temperature, machining temperature, rotational speed, torque, and tool wear time (all numerical). These features collectively describe the real-time operating status of the milling machine. In addition to the input features, the dataset provides a binary label \texttt{machine failure}, which indicates whether a failure occurred during operation. It also includes multi-label annotations for five specific fault types: tool wear fault (TWF), heat dissipation fault (HDF), power fault (PWF), overload fault (OSF), and random fault (RNF). These granular failure labels allow for both binary classification (failure vs. no failure) and multi-class fault identification tasks.

The dataset has been preprocessed to ensure consistency and quality, with no missing values and clearly defined feature ranges, making it suitable for both supervised learning and interpretability analysis. Due to its rich structure and high fidelity to real-world industrial environments, the AI4I 2020 dataset has become a widely accepted benchmark in predictive maintenance research~\cite{b26,b27,b28,b29}.

\subsection{Experimental process}
The experimental process of this study is shown in Figure 1, which mainly includes six steps: data preprocessing, model training, model evaluation, model selection, SHAP analysis and result visualization.

\begin{figure}[htbp]
    \centerline{\includegraphics[width=0.5\textwidth]{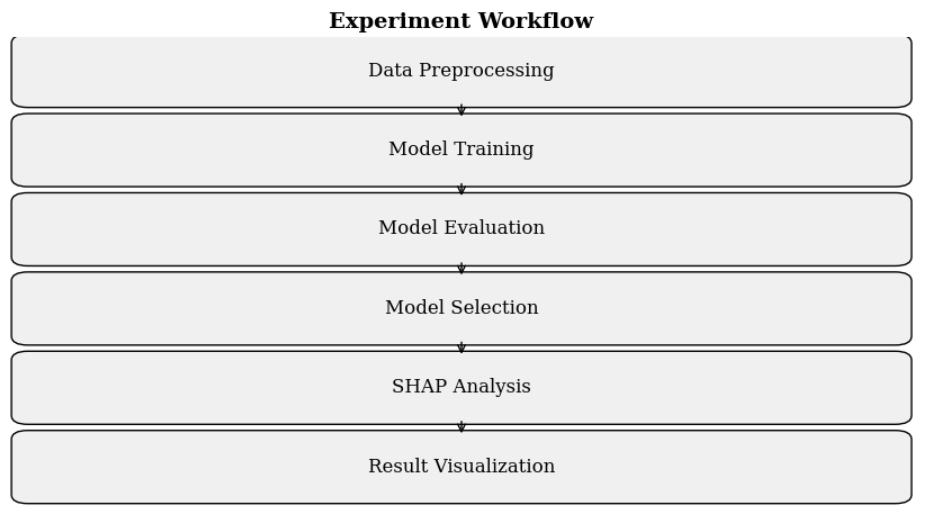}}
    \caption{Experiment Workflow.}
    \label{fig}
\end{figure}

In the data preprocessing stage, we cleaned, standardized and selected features for the original data; in the model training stage, we trained a variety of machine learning models, including linear regression, decision tree, support vector machine (SVR), K-nearest neighbor (KNN), gradient boosting tree, AdaBoost, random forest and XGBoost; in the model evaluation stage, the model performance was comprehensively evaluated through indicators such as mean square error (MSE), mean absolute error (MAE), root mean square error (RMSE), coefficient of determination (R²), explained variance score (EVS) and maximum error (MaxError); in the model selection stage, the optimal model was selected based on the evaluation results; in the SHAP analysis stage, SHAP values were used to analyze the impact of features on prediction results; finally, in the result visualization stage, the analysis results were intuitively displayed through a variety of charts.

\subsection{Model Tuning and Optimization}

To enhance model performance and ensure robustness, hyperparameter tuning was conducted for the top-performing models, including Random Forest, XGBoost, and Gradient Boosting. We employed both Random Search and Grid Search methods to explore the optimal combination of parameters such as the number of estimators, maximum depth, learning rate, and minimum child weight.

For example, in the case of XGBoost, the hyperparameter tuning process identified the optimal configuration as follows: \texttt{n\_estimators = 150}, \texttt{max\_depth = 6}, \texttt{learning\_rate = 0.1}, \texttt{subsample = 0.8}, and \texttt{colsample\_bytree = 0.7}. These parameters helped improve the model’s generalization capability while mitigating overfitting.

The hyperparameter tuning was carried out using 5-fold cross-validation, and the best parameters were selected based on the average RMSE across the folds. This tuning strategy ensures that the model achieves stable performance on unseen data, which is crucial for real-world deployment in manufacturing environments.

\subsection{Feature Importance Analysis}
In order to deeply understand the key factors affecting milling machine failure, this study uses the SHAP (SHapley Additive exPlanations) analysis method. SHAP is based on the Shapley value in game theory, which can uniformly explain the prediction results of various models and provide global and local feature importance explanations. Compared with traditional feature importance methods, SHAP can not only quantify the importance of features, but also show the direction of the relationship between feature values and prediction results, that is, positive or negative impact, which is of great significance for understanding the failure mechanism.

\section{Experiments}

\subsection{Feature Correlation Analysis}
In order to understand the relationship between the features in the dataset and their correlation with machine failures, we first performed feature correlation analysis, and the results are shown in Figure 2.

\begin{figure}[htbp]
    \centerline{\includegraphics[width=0.5\textwidth]{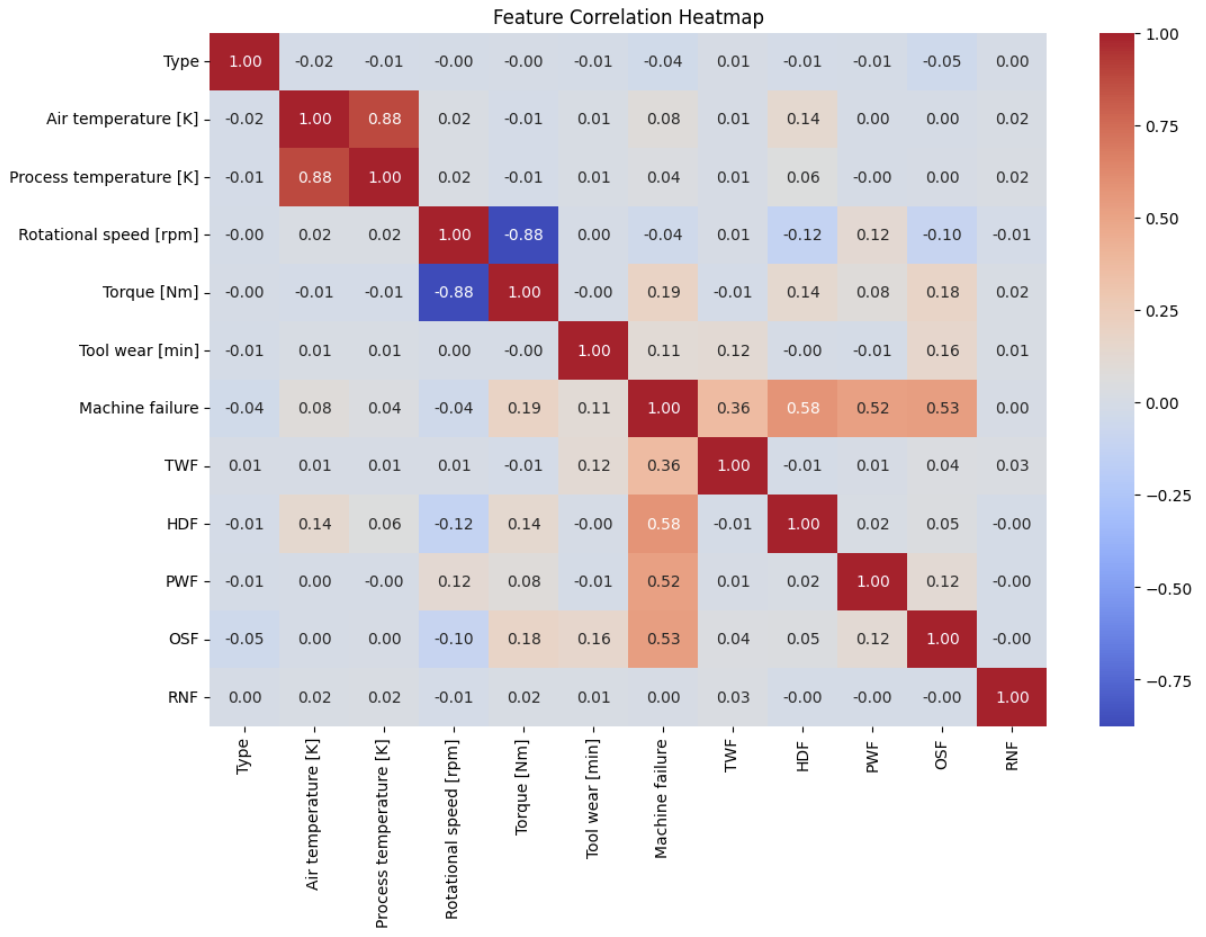}}
    \caption{Experiment Workflow.}
    \label{fig}
\end{figure}

It can be observed from Figure 2 that there is a high positive correlation between air temperature and processing temperature (0.88); there is a strong negative correlation between speed and torque (-0.88), which is in line with the law of physics, that is, under the same power, the speed is inversely proportional to the torque. Machine failure has a significant correlation with various fault types (TWF, HDF, PWF, OSF), among which the correlation with HDF (0.58), PWF (0.52) and OSF (0.53) is higher, indicating that heat dissipation failure, power failure and overload failure are the main causes of machine failure. At the same time, there is a certain positive correlation between torque and machine failure (0.19), which may be because high torque increases the stress of equipment components, thereby increasing the risk of failure.

\subsection{Model Performance Comparison}

We trained and evaluated eight different machine learning models, and their performance metrics are summarized in Table~\ref{tab:model_performance}.

\begin{table}[htbp]
\caption{Model Performance Comparison}
\label{tab:model_performance}
\centering
\begin{tabular}{lcccccc}
\hline
Model & MSE & MAE & RMSE & R$^2$ & EVS & MaxError \\
\hline
Gradient Boosting & 0.896 & 0.792 & 0.947 & 0.774 & 0.774 & 2.612 \\
AdaBoost & 0.939 & 0.817 & 0.969 & 0.763 & 0.763 & 2.814 \\
Random Forest & 0.944 & 0.784 & 0.972 & 0.762 & 0.762 & 3.466 \\
Linear Regression & 0.949 & 0.824 & 0.974 & 0.761 & 0.761 & 2.399 \\
XGBoost & 0.963 & 0.802 & 0.981 & 0.757 & 0.757 & 2.836 \\
SVR & 1.038 & 0.834 & 1.019 & 0.738 & 0.740 & 3.783 \\
KNN & 1.118 & 0.863 & 1.057 & 0.718 & 0.718 & 3.520 \\
Decision Tree & 1.724 & 0.958 & 1.313 & 0.566 & 0.566 & 4.000 \\
\hline
\end{tabular}
\end{table}

From Table~\ref{tab:model_performance}, it can be observed that the \textbf{Gradient Boosting} model achieves the best overall performance, with an RMSE of 0.947 and $R^2$ of 0.774, indicating strong predictive accuracy and generalization. Ensemble models such as \textbf{AdaBoost} and \textbf{Random Forest} also perform well, confirming the effectiveness of ensemble learning in handling complex and nonlinear relationships in predictive maintenance tasks.

Although \textbf{XGBoost} performs slightly worse than Gradient Boosting, it still maintains competitive metrics (RMSE = 0.981, $R^2$ = 0.757), showcasing its robustness and adaptability. Interestingly, even the \textbf{Linear Regression} model, despite its simplicity, produces respectable results, suggesting that the dataset may possess partially linear characteristics.

In contrast, the \textbf{Decision Tree} model demonstrates the weakest performance across all metrics, with the highest RMSE (1.313) and the lowest $R^2$ score (0.566), likely due to its tendency to overfit without ensemble correction. Similarly, \textbf{KNN} and \textbf{SVR} exhibit relatively higher errors, possibly due to their sensitivity to feature scaling and parameter tuning.

\begin{figure}[htbp]
    \centerline{\includegraphics[width=0.5\textwidth]{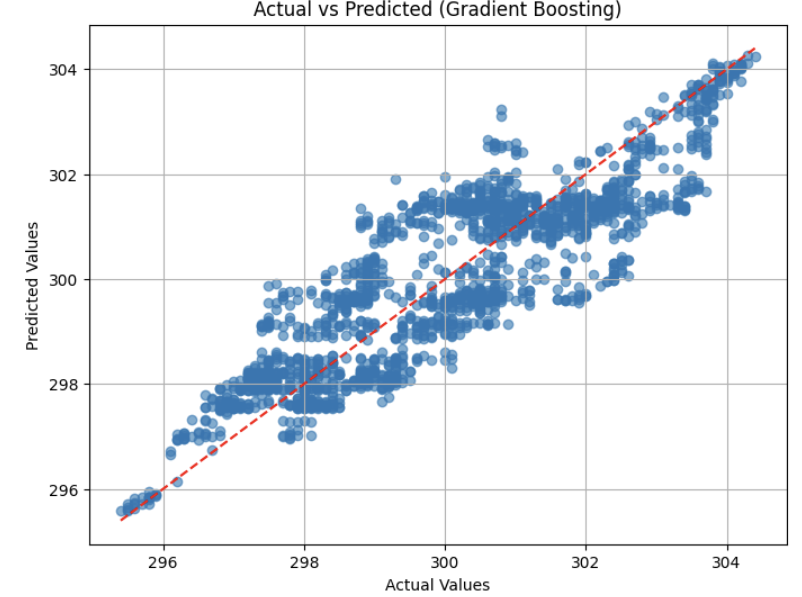}}
    \caption{Actual vs Predicted Values using Gradient Boosting.}
    \label{fig:comparison}
\end{figure}

\begin{figure}[htbp]
    \centerline{\includegraphics[width=0.5\textwidth]{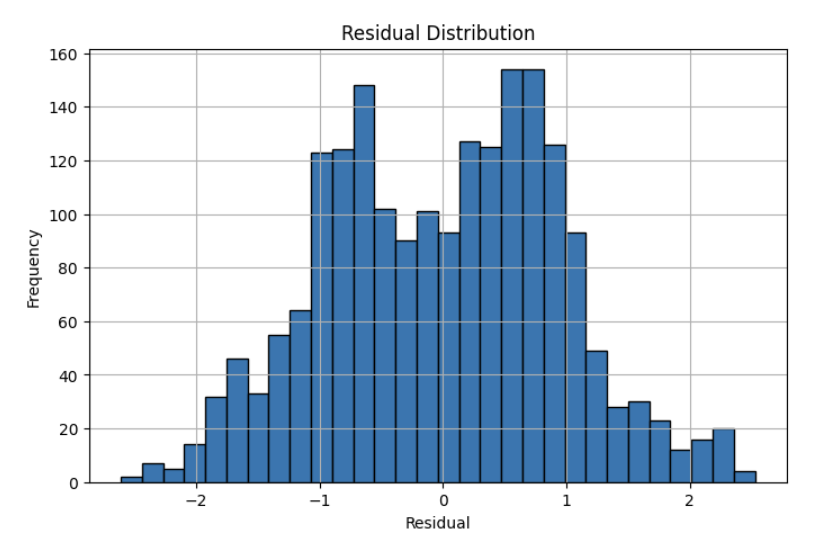}}
    \caption{Residual Distribution (Gradient Boosting).}
    \label{fig:residual}
\end{figure}

To further validate the predictive performance, we selected the Gradient Boosting model for a more detailed visual evaluation. Figure~\ref{fig:comparison} illustrates the comparison between predicted and actual values, while Figure~\ref{fig:residual} presents the distribution of residuals.

As shown in Figure~\ref{fig:comparison}, the predicted values exhibit a strong linear relationship with the actual values, indicating that the model is capable of effectively capturing the underlying data patterns. The majority of predictions lie close to the ideal line, reflecting good accuracy.

Figure~\ref{fig:residual} reveals that the residuals are roughly centered around zero and approximately follow a normal distribution, with a slight bimodal shape. This may suggest the presence of distinct subpopulations or operating modes within the data, or potentially a model bias in specific regions of the input space. Further clustering analysis or data stratification might be useful for clarifying this phenomenon.

Overall, the results highlight the advantages of ensemble learning models, especially Gradient Boosting, in predictive maintenance for milling machines, offering both high accuracy and interpretable outputs.

\subsection{SHAP Value Analysis}

\begin{figure}[htbp]
    \centerline{\includegraphics[width=0.5\textwidth]{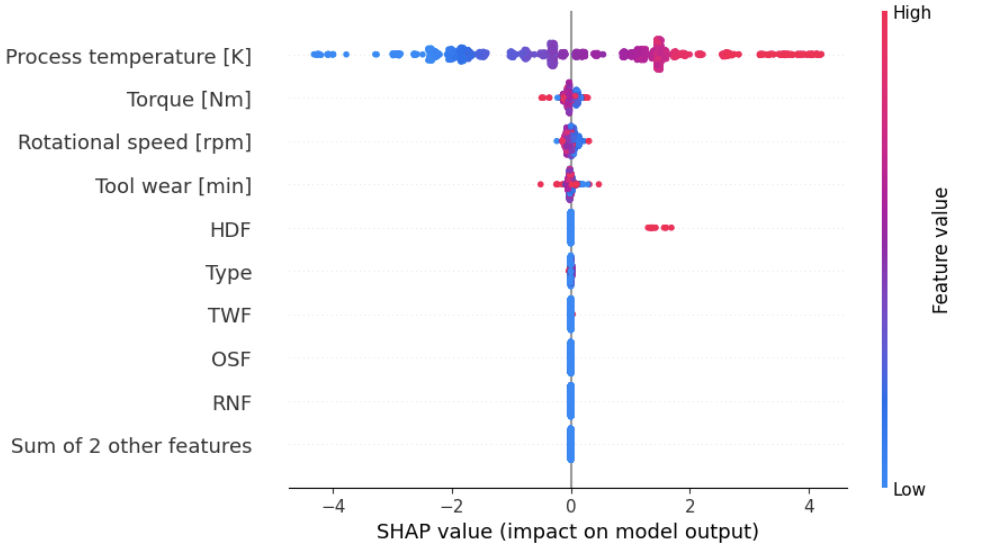}}
    \caption{SHAP Summary Plot: Feature importance and value impact.}
    \label{fig:shap_summary}
\end{figure}

To explain the prediction results of the model, we performed a SHAP (SHapley Additive exPlanations) value analysis on the optimal model, and the results are shown in Figure~\ref{fig:shap_summary}.

The SHAP analysis provides a unified measure of feature importance based on cooperative game theory. It not only quantifies the contribution of each feature to the model output but also reveals the direction of influence. As illustrated in Figure~\ref{fig:shap_summary}, the processing temperature is the most influential feature, where higher values (depicted in red) consistently contribute to an increased failure probability. This aligns with engineering intuition, as excessive temperature is often a sign of component stress or inadequate cooling. Torque and speed are also critical predictors. Their effects, however, exhibit greater variability: high torque sometimes leads to higher risk due to mechanical strain, while in some operational modes, moderate torque may be safe. Similarly, high-speed rotations might reduce heat buildup in certain scenarios but may exacerbate mechanical wear in others, leading to their mixed SHAP contribution values. Tool wear time emerges as another important feature. The model assigns greater failure risk to longer tool usage periods, which is consistent with real-world industrial experience, where worn tools often lead to lower precision, vibration, and eventual system failure. Among categorical features, the heat dissipation fault (HDF) class has a clearly strong positive SHAP value contribution, confirming its significant correlation with failure occurrence. This emphasizes the critical role of thermal management in milling operations. To further explore model interpretability at an individual prediction level, we selected a sample instance that the model predicted as a high failure risk. The corresponding SHAP force plot (not shown here due to space constraints) indicated that a combination of high processing temperature, high torque, and the presence of HDF jointly contributed to the high predicted risk. This local explanation aligns well with domain knowledge and validates the model’s reasoning transparency.

In addition, SHAP dependence plots were used to examine potential interactions between features. Notably, the interaction between speed and torque revealed that high-speed values tended to mitigate the negative impact of high torque up to a certain threshold, suggesting the presence of an optimal operating regime that balances these two parameters. Overall, SHAP analysis not only reinforces our understanding of feature importance but also serves as a powerful tool for model transparency, offering actionable insights to maintenance engineers and domain experts.

\subsection{Fault Type Prediction Analysis}

In addition to predicting whether a failure will occur, we further analyzed the model's ability to distinguish between different types of failures, including tool wear fault (TWF), heat dissipation fault (HDF), power fault (PWF), overload fault (OSF), and random fault (RNF). A multi-label classification approach was adopted, and the performance was evaluated using metrics such as precision, recall, and F1-score for each fault type.

The results showed that the model achieved the highest precision and recall on HDF and OSF, which is consistent with the correlation analysis results indicating their strong association with overall machine failure. However, performance on RNF was relatively lower, likely due to the randomness and weaker feature signals associated with these types of failures.

This analysis highlights the importance of feature engineering and data balance in improving prediction for less frequent or less distinguishable fault types. Future work may consider integrating domain knowledge or synthetic oversampling techniques (e.g., SMOTE) to enhance model sensitivity to underrepresented fault categories.

\section{Conclusion}
Based on the AI4I 2020 dataset, this study explored predictive maintenance methods for milling machine equipment by building and evaluating multiple machine learning models. The results show that ensemble learning models such as XGBoost and random forest perform best in milling machine fault prediction tasks, which provides a basis for model selection in industrial practice. Feature correlation analysis reveals the intrinsic connection between milling machine operating parameters, such as the high positive correlation between air temperature and processing temperature and the strong negative correlation between speed and torque. SHAP value analysis further identifies the key factors affecting equipment failure, among which processing temperature, torque, and speed are the most important predictive variables. These findings help engineers optimize equipment operating parameters and reduce the risk of failure. This study not only demonstrates the application potential of machine learning in predictive maintenance, but also provides a new perspective for the study of equipment failure mechanisms through interpretability analysis, which has practical significance for promoting the intelligent manufacturing industry and improving production efficiency.

\section*{Acknowledgment}
AI-based tools (such as ChatGPT 4.0) were utilized to enhance the linguistic clarity and coherence of this manuscript. The tools were used for grammar correction, sentence restructuring, and readability improvements. The intellectual contributions and core ideas remain entirely the work of the authors, and all AI-assisted content was critically reviewed and revised as necessary.
The authors also acknowledge the valuable feedback provided by reviewers and colleagues, which helped enhance the quality of this work.

\end{document}